\pdfoutput=1

\documentclass[11pt]{article}

\usepackage[]{acl}

\usepackage{times}
\usepackage{latexsym}

\usepackage[T1]{fontenc}

\usepackage[utf8]{inputenc}

\usepackage{microtype}
\usepackage{booktabs}
\usepackage{amsmath}
\usepackage{multirow}
\usepackage{color}
\usepackage{footmisc}
\usepackage{soul}
\usepackage{arydshln}
\usepackage{graphics}
\usepackage{graphicx}
\usepackage{algorithm}

\usepackage{float}
\usepackage{subfigure}
\usepackage{caption}
\usepackage{algpseudocode}
\definecolor{lightgreen}{HTML}{CEF6CE}
\definecolor{lightred}{HTML}{FD8F70}

\usepackage{amsfonts}

\title{Disentangling Confidence Score Distribution for Out-of-Domain Intent Detection with Energy-Based Learning}

\author{Yanan Wu$^{1*}$, Zhiyuan Zeng$^{1*}$, Keqing He$^{2*}$, Yutao Mou$^{1}$ ,\\ \bf{Pei Wang$^{1}$}, \bf {Yuanmeng Yan$^{1}$}, {\bf Weiran Xu$^{1}$}\thanks{\ \ The first three authors contribute equally. Weiran Xu is the corresponding author.}\\
  $^1$Beijing University of Posts and Telecommunications, Beijing, China\\
$^{2}$Meituan, Beijing, China\\
  \texttt{\{yanan.wu,zengzhiyuan,myt,wangpei,}\\ \texttt{yanyuanmeng,xuweiran\}@bupt.edu.cn}\\
  \texttt{\{hekeqing\}@meituan.com}
  }
  
\begin{document}
\maketitle
\begin{abstract}
Detecting Out-of-Domain (OOD) or unknown intents from user queries is essential in a task-oriented dialog system. Traditional softmax-based confidence scores are susceptible to the overconfidence issue. In this paper, we propose a simple but strong energy-based score function to detect OOD where the energy scores of OOD samples are higher than IND samples. Further, given a small set of labeled OOD samples, we introduce an energy-based margin objective for supervised OOD detection to explicitly distinguish OOD samples from INDs. Comprehensive experiments and analysis prove our method helps disentangle confidence score distributions of IND and OOD data.\footnote{Our code is available at \url{https://github.com/pris-nlp/EMNLP2022-energy_for_OOD/}.}
\end{abstract}

\vspace{0.13cm}
\section{Introduction}

Detecting Out-of-Domain (OOD) or unknown intents from user queries is crucial to a task-oriented dialog system \cite{Akasaki2017ChatDI,Tulshan2018SurveyOV,Shum2018FromET,Lin2019DeepUI,xu-etal-2020-deep,zeng-etal-2021-modeling,Wu2022DistributionCF}. It can avoid performing wrong operations and provide potential directions of future development when an input query falls outside the range of predefined intents. Since the exact number of unknown intents in practical scenarios is hard to know and annotate, the lack of real OOD examples makes it challenging to identify these samples in dialog systems.

\begin{figure}[t]
    \centering
    \resizebox{.485\textwidth}{!}{
    \includegraphics{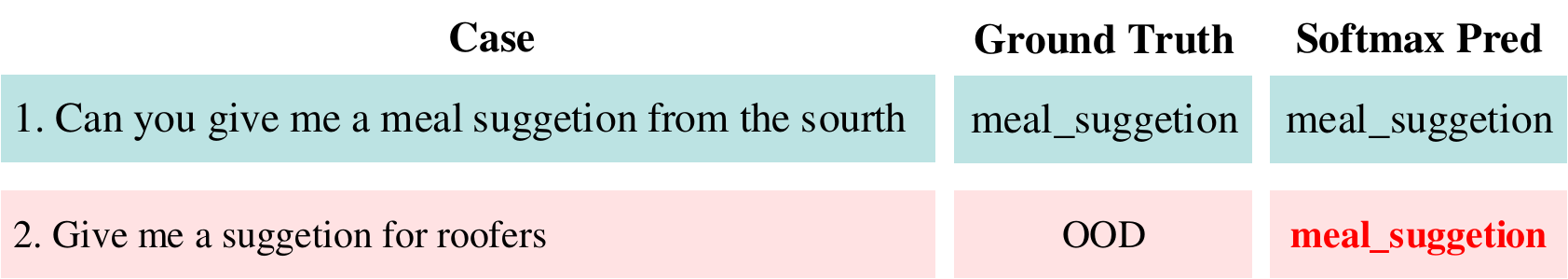}}
    \vspace{-0.6cm}
    \caption{IND(case 1) vs OOD sample(case 2). Softmax score recognizes OOD sample as IND intent type because of overconfidence issue.}
    \label{case1}
     \vspace{-0.3cm}
\end{figure}

\begin{figure}[t]
    \centering
    \resizebox{.48\textwidth}{!}{
    \includegraphics{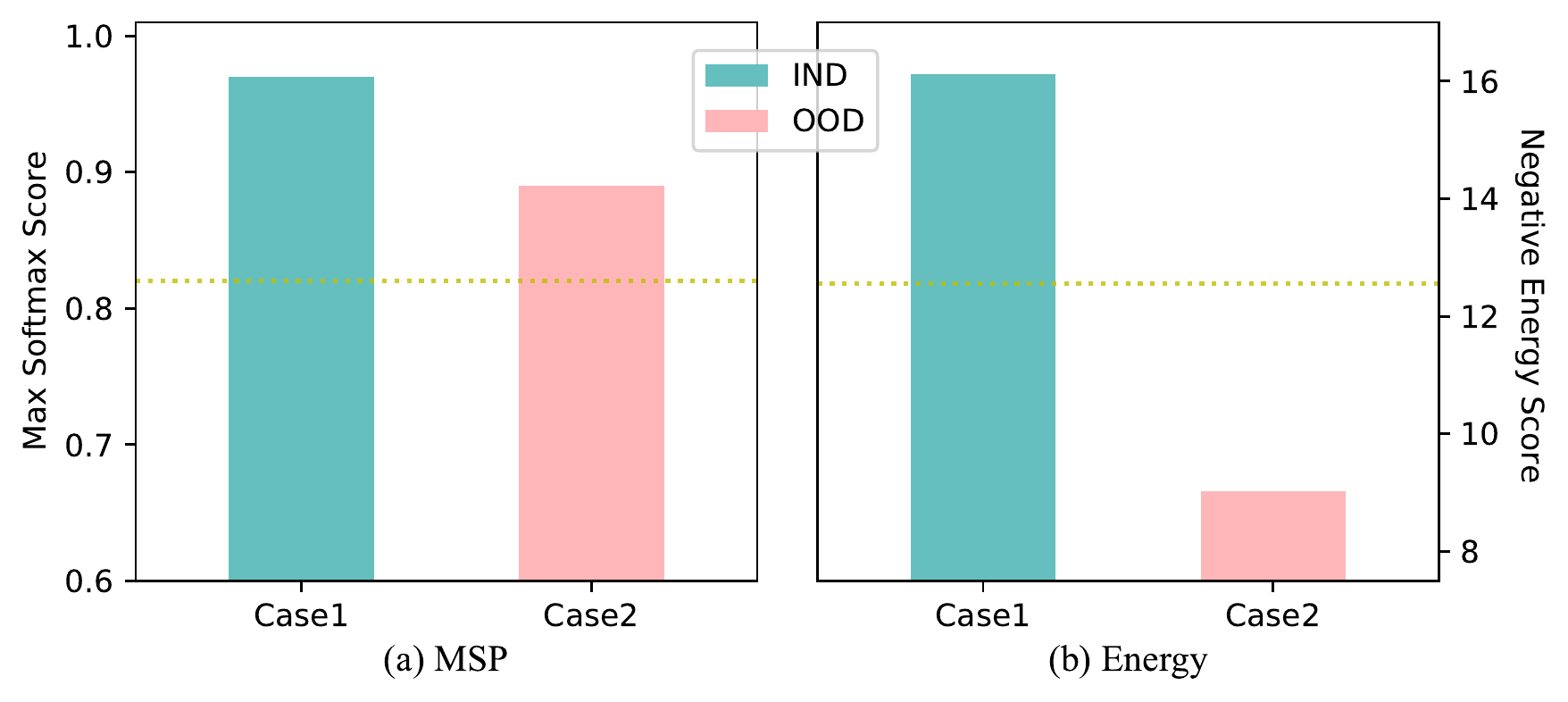}}
    \vspace{-0.7cm}
    \caption{Softmax score from MSP vs energy score from our method. Softmax score are similar for IND and OOD (both $>0.85$) but energy score are more distinguished.}
    \label{case2}
    \vspace{-0.2cm}
\end{figure}

Depending on whether labeled OOD samples are available, previous OOD detection work can be generally classified into two types: unsupervised  \cite{Bendale2016TowardsOS,Hendrycks2017ABF,Shu2017DOCDO,Lee2018ASU,Ren2019LikelihoodRF,Lin2019DeepUI,xu-etal-2020-deep,zeng-etal-2021-modeling,Zeng2021AdversarialSL,Wu2022RevisitOF} and supervised \cite{Fei2016BreakingTC,Kim2018JointLO,Larson2019AnED,Zheng2020OutofDomainDF}. The former firstly learn an in-domain (IND) intent classifier only using labeled IND data and then estimates the confidence score of a test query. For example, Maximum Softmax Probability (MSP) \cite{Hendrycks2017ABF} uses maximum softmax probability as the confidence score and regards an intent as OOD if the score is below a fixed threshold. The assumption is that OOD intents should produce a lower softmax probability than INDs. However, neural networks can produce arbitrarily high softmax confidence even for such abnormal OOD samples \cite{Guo2017OnCO,Liang2018EnhancingTR}, as shown in Fig \ref{case1}\&\ref{case2}, which we call \emph{overconfidence}. 
\begin{figure*}[t]
    \centering
    \resizebox{0.9\linewidth}{!}{
    \includegraphics{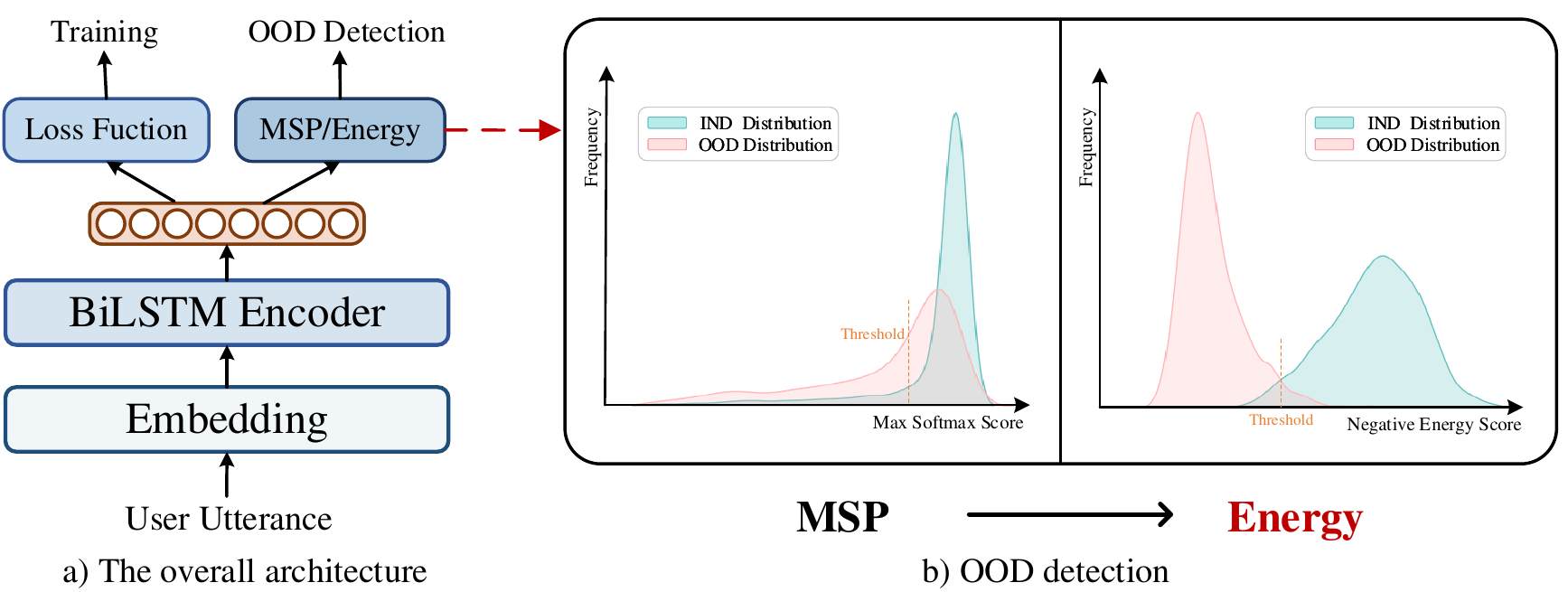}}
    \vspace{-0.3cm}
    \caption{The overall architecture of our proposed method.}
    \label{model}
     \vspace{-0.4cm}
\end{figure*}
Further, another distance-based method, Gaussian discriminant analysis (GDA)  \cite{xu-etal-2020-deep}, is proposed to use the maximum Mahalanobis distance \cite{mahalanobis1936generalized} to all in-domain classes centroids as the confidence score. Compared to MSP, GDA gets better OOD performance but requires expensive computation for complex Mahalanobis distance. In this paper, we aim to use simple softmax confidence scores for both higher performance and efficiency. For supervised OOD detection, \newcite{Fei2016BreakingTC,Larson2019AnED}, form a \emph{(N+1)}-class classification problem where the \emph{(N+1)}-th class represents the OOD intents. Further, \newcite{Zheng2020OutofDomainDF} uses labeled OOD data to generate an entropy regularization term. But these methods require numerous labeled OOD intents to get superior results. We focus on using fewer labeled OOD data (like 20 or 30) to achieve comparable even better performance.

In this paper, we propose an energy-based score function to detect OOD in an unsupervised manner. The energy-based score function maps each query to a single energy scalar which is lower for IND samples and higher for OOD samples based on the energy theory \cite{LeCun2006ATO}. We first train an in-domain intent classifier via IND data, then replace the original softmax layer with the energy-based score function. Our method can not only mitigate the issue of overconfident softmax probability but also reduce expensive post-processing computation. Further, given a small portion of labeled OOD samples, we propose an energy-based margin objective to explicitly distinguish OOD samples from IND samples. Our contributions are three-fold: (1) We propose an energy-based learning method for OOD intent detection to achieve higher performance and efficiency. (2) We propose an energy-based margin objective to distinguish energy distributions of OOD and IND samples. (3) Extensive experiments and analysis on two benchmarks demonstrate the effectiveness of our method.

\begin{table*}[t]
\centering
\resizebox{0.82\textwidth}{!}{%
\begin{tabular}{c|l|cc|cc|cc|cc}
\hline
\multicolumn{2}{c|}{\multirow{3}{*}{Models}} & \multicolumn{4}{c|}{CLINC-Full} & \multicolumn{4}{c}{CLINC-Small} \\ \cline{3-10} 
\multicolumn{2}{c|}{} & \multicolumn{2}{c|}{IND} & \multicolumn{2}{c|}{OOD} & \multicolumn{2}{c|}{IND} & \multicolumn{2}{c}{OOD} \\ \cline{3-10} 
\multicolumn{2}{c|}{} & Acc & F1 & Recall & F1 & Acc & F1 & Recall & F1 \\ \hline
\multicolumn{1}{c|}{\multirow{4}{*}{\begin{tabular}[c]{@{}c@{}}Unsupervised \\ OOD\end{tabular}}} 
																	&  MSP  \cite{Hendrycks2017ABF}                       & 87.16           & 87.64           & 41.40           & 44.86           & 85.02 & 85.18 & 35.81 & 36.60                       \\
                                  & LOF \cite{Lin2019DeepUI}                        & 85.87           & 86.08           & 58.32           & 59.28           &  82.83 & 82.98 & 53.96 & 54.63 \\
                                  & GDA \cite{xu-etal-2020-deep}                        & 86.83           & 87.90           & 64.14           & 65.79           & 84.46 & 84.87 & 60.72 & 61.89                        \\
                                          &SCL\cite{zeng-etal-2021-modeling} & 87.01  & 88.28  & 66.80  & 67.68  & 85.73  & 86.61  & 63.96  & 64.44 \\
                                  & Energy(Ours)                & \textbf{88.71}           & \textbf{89.17}           & \textbf{68.10}           & \textbf{69.64}           &\textbf{86.42} & \textbf{86.48} & \textbf{65.78} & \textbf{66.52}                \\ \hline \hline
\multirow{10}{*}{\begin{tabular}[c]{@{}c@{}}Supervised \\ OOD\end{tabular}}  & N+1                         & \textbf{91.24} & 85.29           & 24.51           & 31.08           &    \textbf{90.13} & 83.23 & 21.50 & 29.17              \\ \cline{2-10}
                                  & MSP+Entropy \cite{Zheng2020OutofDomainDF}                 & 87.48           & 87.81           & 49.90           & 53.93           &     85.24 & 85.31 & 45.90 & 48.57      \\
                                   & MSP+Bound \cite{liu2020energy} & 88.03 & 87.26 & 45.21 & 56.86 & 86.16 & 83.04 & 42.38 & 51.43 \\ & MSP+Margin(Ours)                  & 88.31           & 87.98           & \textbf{57.27}           & \textbf{59.96}           &        85.33 & 85.37 & \textbf{54.90} & \textbf{55.37}      \\ \cline{2-10}
                                  & LOF+Entropy                 & 85.98           & 86.37           & 61.10           & 61.13       &   83.49 & 83.86 & 57.70 & 57.79    \\
                                   & LOF+Bound & 86.36 & 85.66 & 57.83 & 60.15 & 81.36 & 82.88 & 64.41 & 59.30 \\ 
                                   & LOF+Margin(Ours)                  & 86.13           & 86.59           & \textbf{65.70}           & \textbf{65.59}           &       83.57 & 83.97 & 63.60 & \textbf{63.18}                    \\ \cline{2-10}
                                  & GDA+Entropy                 & 87.27           & 88.14           & 68.53           & 68.82           &    85.01 & 85.53 & 65.22 & 65.65   \\
                                  & GDA+Bound & 87.09 & 86.86 & 67.32 & 66.41 & 84.44 & 84.75 & 65.19 & 64.14 \\ 
                                  & GDA+Margin(Ours)                  & 87.54           & 88.23           & 68.42           & 68.73           &        85.51 & 85.81 & 65.13 & 65.68                   \\ \cline{2-10}
                                
                                \rule{0pt}{12pt}  & Energy+Margin(Ours, Full Model)         & \textbf{89.75}           & \textbf{89.46}  & \textbf{73.92}  & \textbf{74.06}  &        \textbf{87.84} & \textbf{87.53} & \textbf{72.76} & \textbf{72.98}          \\
                                  
                                  \hline
\end{tabular}}
\vspace{-0.2cm}
\caption{Performance comparison on CLINC-Full and CLINC-Small datasets ($p < 0.01$ under t-test).}
\label{main_result}
\end{table*}

\vspace{0.1cm}
\section{Methodology}

\textbf{Overall Architecture} Fig \ref{model}(a) shows the overall architecture of our proposed method. We first train an in-domain intent classifier using IND data in training stage. Then in the test stage, we extract the intent feature of a test query and employ the detection algorithms MSP \cite{Hendrycks2017ABF} or Energy to detect OOD. Fig \ref{model}(b) demonstrates the effectiveness of our method distinguishing OOD distributions from IND\footnote{Because the max softmax score is higher for IND samples and lower for OOD samples, we use the negative energy score to align with the conventional definition where positive(IND) samples get higher scores.}.

\textbf{Energy-based Score Function} To mitigate the issue of overconfident softmax probability in MSP, we propose an energy-based score function to push apart score distributions of OOD and IND samples. We first briefly review the energy theory \cite{LeCun2006ATO} then explain our proposed energy-based score function for OOD detection. The previous energy work \cite{LeCun2006ATO,Zhai2016DeepSE,Grathwohl2020YourCI,Liu2020EnergybasedOD,Kaur2021AreAO} aims to build a function $E(\mathbf{x}): {R}^{D} \rightarrow {R}$ which maps a sample $\mathbf{x}$ to a single scalar called the \emph{energy}. Given a data point $\mathbf{x} \in {R}^{D}$, the energy function can be defined as follows:

\begin{equation}
\setlength{\abovedisplayskip}{0.05cm}
\setlength{\belowdisplayskip}{0.00cm}
E(\mathbf{x})=-T \cdot \log \int_{y^{\prime}} e^{-E\left(\mathbf{x}, y^{\prime}\right) / T}
\label{eq1}
\end{equation}
where $T$ is the temperature parameter and $E(\mathbf{x}, y^{\prime})$ is the marginal energy over label $y^{\prime}$. Essentially, energy scores can be transfered to the likelihood probability: 

\begin{equation}
\setlength{\abovedisplayskip}{0.05cm}
\setlength{\belowdisplayskip}{0.4cm}
p(y \mid \mathbf{x})=\frac{e^{-E(\mathbf{x}, y) / T}}{\int_{y^{\prime}} e^{-E\left(\mathbf{x}, y^{\prime}\right) / T}}=\frac{e^{-E(\mathbf{x}, y) / T}}{e^{-E(\mathbf{x}) / T}}
\end{equation}
For OOD detection, since we focus on the detection algorithms for the test stage in this paper, we train the same BiLSTM in-domain intent classifier $f(\mathbf{x})$ via IND data as \newcite{Lin2019DeepUI} in the training stage. Then given a test query, we simply use the logits from the intent classifier to represent $E(\mathbf{x}, y^{\prime})$. Therefore, the energy score function Eq \ref{eq1} can be formulated as:

\begin{equation}
E(\mathbf{x} ; f)=-T \cdot \log \sum_{i}^{K} e^{f_{i}(\mathbf{x}) / T}
\label{eq3}
\end{equation}
where $K$ is the size of IND intent classes and $f_{i}(\mathbf{x})$ is the logit of $\mathbf{x}$ belonging to $i$-th class. 
We simply use a threshold on the energy score to consider whether a test query belongs to OOD. 
Intuitively, the reason why the energy score works for OOD detection is that higher energy represents a lower likelihood of occurrence according to \newcite{LeCun2006ATO}. Therefore, unobserved OOD samples in the training stage should get lower likelihoods as well as higher energy scores than observed IND samples. In Appendix \ref{proof}, we provide a detailed theoretical derivation of why the energy function can alleviate the overconfidence problem. Besides, Experiment \ref{Distribution_analysis} also proves energy scores better distinguish confidence distribution of OOD data from IND data than softmax probabilities.

\textbf{Energy-guided Margin Objective} To further distinguish OOD from IND, we propose an energy-guided margin objective for few-shot supervised OOD detection. Different from \citet{liu2020energy}, our approach directly models the energy boundary by pushing apart the samples from IND and OOD, which helps recognize OOD intents near the decision boundary and is easier to tune and less sensitive to the noise. Specifically, we use an energy-based max-margin loss as well as the standard cross-entropy loss to explicitly set an energy gap between OOD and IND. We aim to learn more discriminative representations for energy score distributions in the training stage. The energy margin loss is formulated as:
\begin{table}[t]
\centering
\resizebox{0.48\textwidth}{!}{%
\begin{tabular}{l|cc}
\hline
CLINC               & Full       & Small \\ \hline
Avg utterance length & 9          & 9           \\
Intents              & 150        & 150         \\
Training set size    & 15100      & 7600        \\
Training samples per class  & 100      & 50 \\
Training OOD samples amount & 100      & 100 \\
Development set size & 3100       & 3100        \\
Development samples per class & 20       & 20        \\
Development OOD samples amount & 100       & 100 \\
Testing Set Size     & 5500       & 5500        \\
Testing samples per class & 30       & 30        \\
Development OOD samples amount & 1000       & 1000 \\ \hline
\end{tabular}
}
\caption{Statistics of the CLINC datasets.}
\label{dataset}
\end{table}

\begin{align}
\setlength{\abovedisplayskip}{0.1cm}
\setlength{\belowdisplayskip}{0.00cm}
\mathcal{L}\!=\!\mathbb{E}_{\left(\mathbf{x}_{\text {ind}},\mathbf{x}_{\text {ood}}\right) \sim \mathcal{D}}\max \!(0, m\!+ \!E(\mathbf{x}_{\text{ind}})\!-\!E(\mathbf{x}_{\text{ood}}))
\end{align}

where $m$ is the energy margin and $E$ is the energy score of IND or OOD samples in train set. Then in the test stage, we still use the energy score to detect OOD. Analysis \ref{Unsupervised_vs_supervised} displays the effectiveness of the margin loss over unsupervised OOD.

\vspace{0.2cm}
\section{Experiments}
\vspace{0.2cm}
\subsection{Datasets}
\begin{figure*}[t]
    \centering
    \resizebox{0.9\linewidth}{!}{
    \includegraphics{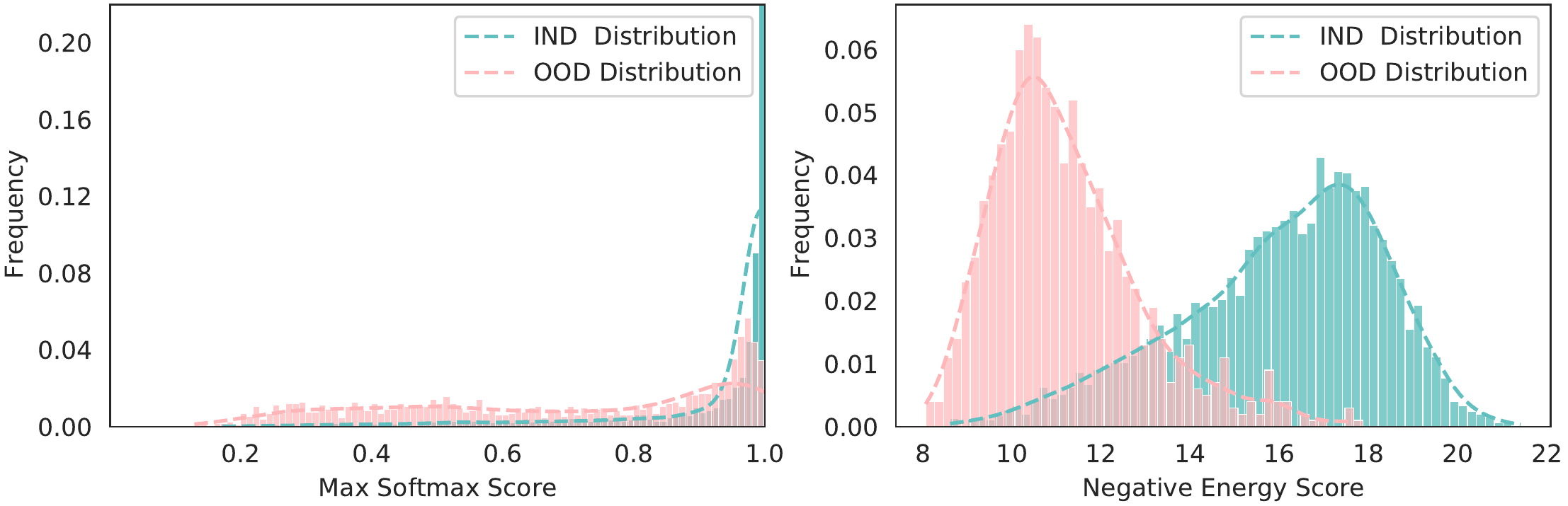}}
    \caption{Distribution of softmax scores vs energy scores.}
    \label{analysis_1}
\end{figure*}

We use two public benchmark OOD datasets\footnote{https://github.com/clinc/oos-eval}, CLINC-Full and CLINC-Small\cite{larson-etal-2019-evaluation}. We show the detailed statistic of these datasets in Table \ref{dataset}. They both contain 150 in-domain intents across 10 domains. The difference is that CLINC-Small has fewer in-domain training examples than CLINC-Full. Note that all the datasets we used have a fixed set of labeled OOD data but we don't use it for training.

\subsection{Metrics} 
We report both OOD metrics: Recall and F1-score(F1) and in-domain metrics: F1-score(F1) and Accuracy(ACC). Since we aim to improve the performance of detecting out-of-domain intents from user queries, OOD Recall and F1 are the main evaluation metrics in this paper.

\subsection{Baselines} 
For detection algorithms, we use MSP, LOF and GDA as baselines. For training objectives, we use N+1, entropy and bound as baselines. We present dataset statistics, baselines and implementation details in the appendix. We will release our code after blind review.

\subsection{Main Results}
Table \ref{main_result} shows the main results. 
(1) For unsupervised OOD detection, using the energy function achieves 24.78, 10.36, 3.85, 1.96 OOD F1 improvements over MSP, LOF, GDA and SCL on CLINC-Full. The results prove the effectiveness of energy score function for OOD detection. Besides, for IND metrics, energy function also outperforms SCL by 0.89\%(F1), which reflects energy scores can better distinguish OOD from IND samples without sacrificing IND performance. (2) For supervised OOD detection, we compare different pre-training losses under the same detection score function. We find our Margin approach achieve consistent improvements under different detection functions on both datasets. It demonstrates that Margin objective can stably improve the representation space by directly pushing apart the samples from IND and OOD. We also observe under MSP, our proposed Margin objective outperforms Entropy by 6.03\% and Bound by 3.10\% on CLINC-Full. But on GDA we find no significant performance difference. We argue the energy-based learning may not always fit in generative distance-based detection methods like GDA. Overall, combining energy score function and margin objective achieve the best performance over the previous state-of-the-art by 5.24\%.
\begin{figure}[t]
    \centering
    \resizebox{0.90\linewidth}{!}{
    \includegraphics{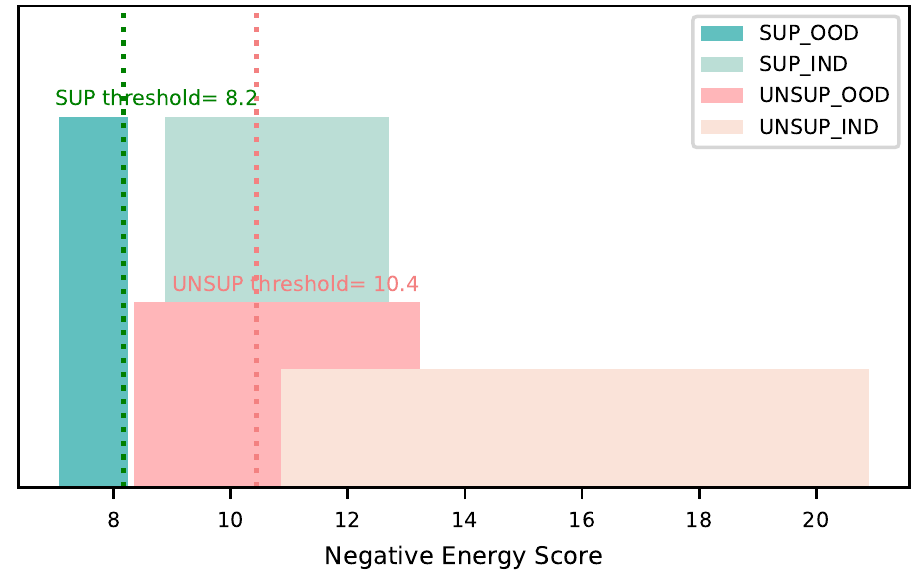}}
    \caption{Unsupervised vs supervised OOD detection.}
    \label{analysis_2}
\end{figure}

\vspace{0.2cm}
\section{Analysis}
\vspace{0.2cm}

\subsection{Distribution of softmax scores vs energy scores}
\label{Distribution_analysis}
To figure out why energy scores outperform softmax scores, we compare the score histogram distributions for IND and OOD data in Fig \ref{analysis_1}. We use the same pre-trained intent classifier to compute scores on the test set. We find softmax scores for both IND and OOD data concentrate on high values, resulting in severe overconfidence. By contrast, energy scores better distinguish score distribution of OOD data from IND data. And energy distributions are smoother than softmax score distributions. Overall, our proposed energy-based score function can disentangle confidence score distributions for IND and OOD data.

\label{Unsupervised_vs_supervised}
\subsection{Unsupervised vs supervised OOD detection} 
To verify the effectiveness of our proposed energy-based margin objective, we compare the energy score statistics of unsupervised (Energy) and supervised (Margin+Energy) OOD detection in Fig \ref{analysis_2}. Each rectangle in Fig \ref{analysis_2} represents the energy distribution of IND or OOD data, where the middle of the rectangle is energy mean and the width of the rectangle is energy variance. Results show that compared to Energy, Margin+Energy makes negative energy scores of both OOD and IND data smaller. Further, the supervised Margin objective can significantly decrease the variance of both OOD ($1.86\downarrow$) and IND ($3.11\downarrow$) data. Therefore, Margin can push apart energy score distributions for OOD detection by shrinking its variance to avoid overlapping. Besides, combined with the energy threshold (dot line in Fig \ref{analysis_2}), unsupervised (Energy) still gets a portion of OOD samples above the threshold which are misclassified into IND, but supervised (Margin+Energy) on the opposite. It proves that Margin can further mitigate the issue of overconfidence.

\begin{figure}[t]
    \centering
    \resizebox{0.90\linewidth}{!}{
    \includegraphics{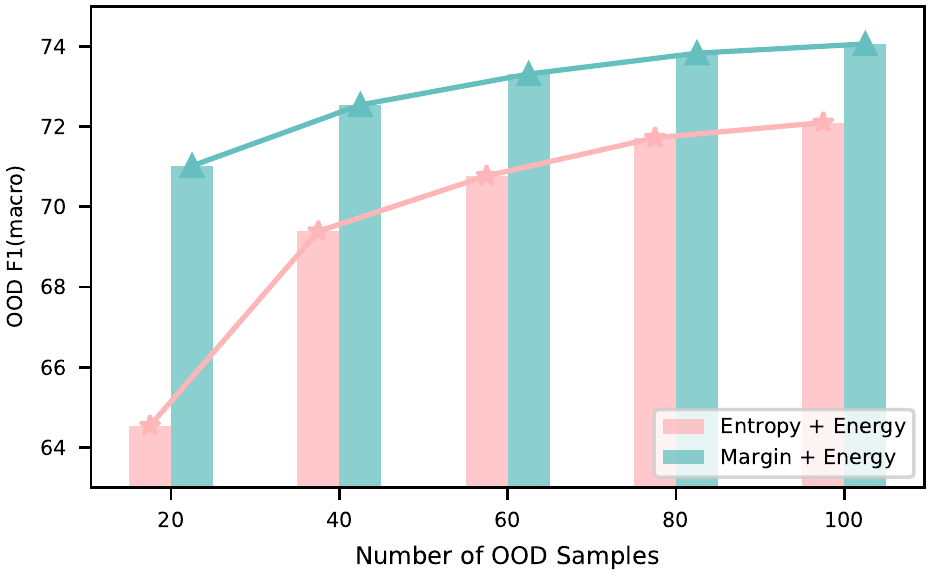}}
    \caption{Effect of number of labeled OOD samples.}
    \label{analysis_3}
\end{figure}

\subsection{Effect of number of labeled OOD samples} 
Fig \ref{analysis_3} shows the effect of labeled OOD training data size for supervised OOD detection. We find Margin+Energy consistently outperforms Entropy+Energy, especially in the few-shot supervised OOD scenario, which demonstrates strong robustness and generalization of our proposed energy-based margin objective for OOD detection.

\subsection{Effect of Parameters}
\begin{figure}[t]
    \centering
    \resizebox{0.9\linewidth}{!}{
    \includegraphics{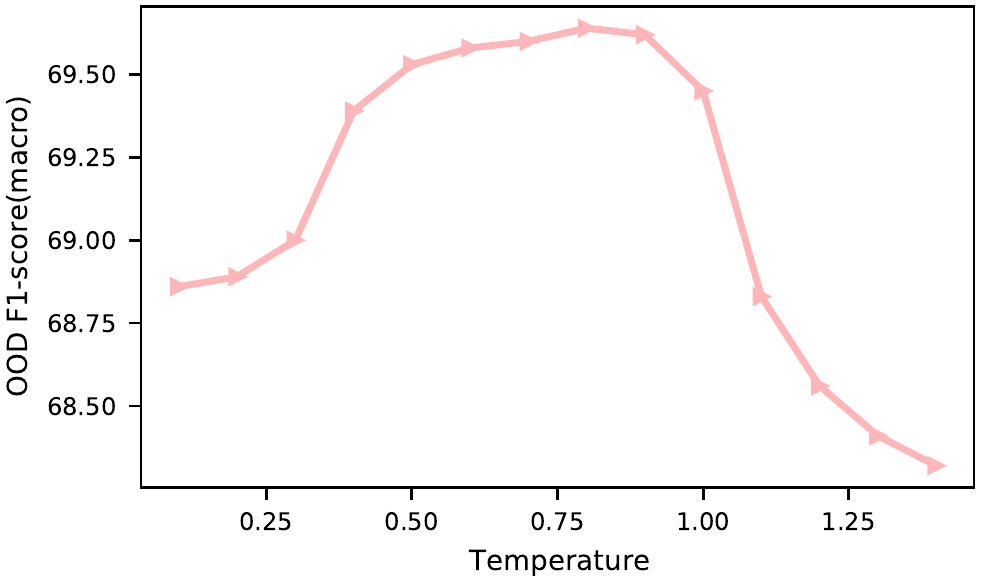}}
    \caption{Effect of energy temperature $T$}
    \label{analysis_temperature}
\end{figure}

\textbf{Temperature $T$.} Fig \ref{analysis_temperature} shows the effect of different energy temperature $T$. We conduct the experiments on the CLINC-Full dataset, using Energy for unsupervised OOD. The X-axis denotes the value of temperature $T$. In general, $T \in (0.5, 1.0)$ achieves relatively better performances and has a broad range. 

\begin{figure}[t]
    \centering
    \resizebox{0.9\linewidth}{!}{
    \includegraphics{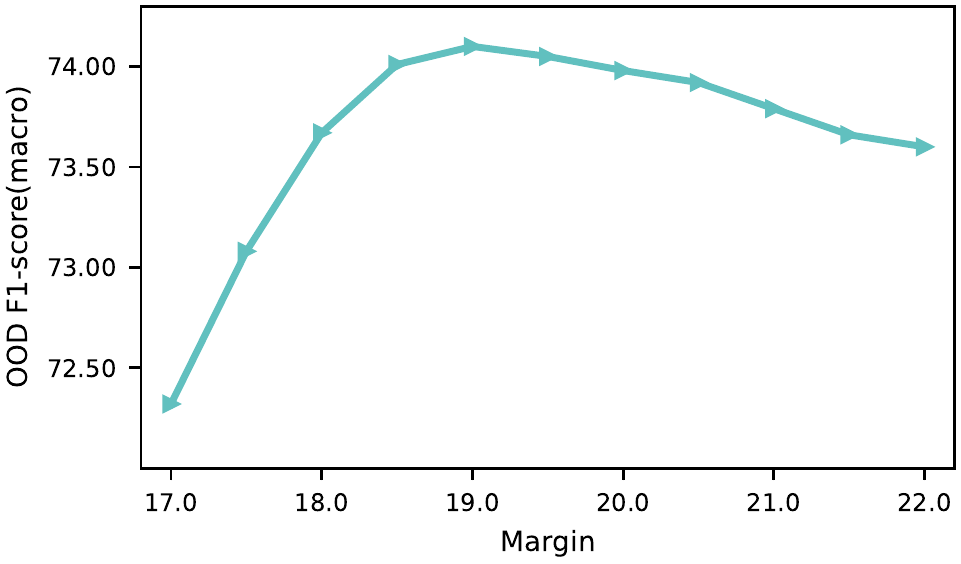}}
    \caption{Effect of energy margin $m$}
    \label{analysis_margin}
\end{figure}

\textbf{Margin $m$.} Fig \ref{analysis_margin} shows the effect of different energy margin $m$. We conduct the experiments on the CLINC-Full dataset, using Margin+Energy for supervised OOD. The X-axis denotes the value of margin $m$. Results show that $m = 19.0$ achieves the best performance and is robust to minor changes.

\section{Conclusion}
Traditional softmax-based OOD detection methods 
are susceptible to the overconfidence issue. Therefore, we propose a novel energy-based score function to mitigate the issue of softmax overconfidence. To use labeled OOD data, we further introduce an energy-based margin objective to explicitly distinguish energy score distributions of OOD from IND. Experiments and analysis confirm the effectiveness of our energy-based method for OOD detection. For future work, we hope to explore theoretical concepts of energy and provide new guidance.


\bibliography{anthology,custom}
\bibliographystyle{acl_natbib}

\appendix




\section{Baseline Details}
We perform main experiments based on two different settings, unsupervised OOD and supervised OOD detection. For unsupervised OOD detection, we compare our proposed energy detection algorithm with other methods, MSP (Maximum Softmax Probability)\cite{Hendrycks2017ABF}, LOF (Local Outlier Factor)\cite{Lin2019DeepUI}, GDA (Gaussian Discriminant Analysis)\cite{xu-etal-2020-deep}.  For supervised OOD detection, we also compare our proposed energy-based margin objective with entropy \cite{Zheng2020OutofDomainDF} and N+1 \cite{Fei2016BreakingTC,Larson2019AnED}. Note that margin and entropy objectives are used in the training stage, we still need detection algorithms MSP, GDA or Energy to detect in the test stage. We supplement the relevant baseline details as follows: 

\noindent\textbf{MSP} (Maximum Softmax Probability)\cite{Hendrycks2017ABF} uses maximum softmax probability as the confidence score and regards an intent as OOD if the score is below a fixed threshold.

\noindent\textbf{LOF} (Local Outlier Factor)\cite{Lin2019DeepUI} uses the local outlier factor to detect unknown intents. The motivation is that if an example’s local density is significantly lower than its k-nearest neighbor’s, it is more likely to be considered as the unknown intents.

\noindent\textbf{GDA} (Gaussian Discriminant Analysis)\cite{xu-etal-2020-deep} is a generative distance-based classifier for out-of-domain detection with Euclidean space. They estimate the class-conditional distribution on feature spaces of DNNs via Gaussian discriminant analysis (GDA) to avoid over-confidence problems and use Mahalanobis distance to measure the confidence score of whether a test sample belongs to OOD. GDA is the state-of-the-art detection method till now, our proposed energy score still significantly outperforms GDA. 

Note that LOF and GDA both require additional post-processing modules to estimate density or distance, which induces expensive computation. We conduct a performance comparison for inference time in Table \ref{time}. Since SCL only adds a pre-training loss along with CE and also uses GDA for detection, the inference time is equal to GDA.

\begin{table}[h]
\centering
\resizebox{0.3\textwidth}{!}{%
\begin{tabular}{l|cc}
\hline
Detect    Method           & Inference   time   \\ \hline
MSP &  1.00x \\\hline
Energy(Ours) & 1.00x \\  \hline
GDA/SCL              &       30.63x   \\ \hline
LOF    &    30.89x \\ \hline
\end{tabular}
}
\caption{Inference time comparison between different methods.}
\label{time}
\end{table}

\noindent\textbf{SCL}\cite{zeng-etal-2021-modeling} uses a supervised contrastive learning objective to minimize intra-class variance by pulling together in-domain intents belonging to the same class and maximize inter-class variance by pushing apart samples from different classes. Note that SCL still needs a confidence score function. To keep fair comparison, we follow the original paper using GDA detection method.

\noindent\textbf{N+1}(\newcite{Fei2016BreakingTC,Larson2019AnED}) is an N+1 classification model which simply considers OOD samples as a new class. 

\noindent\textbf{Entropy}(\newcite{Zheng2020OutofDomainDF}) uses labeled OOD data to generate an entropy regularization term to enforce the predicted distribution of OOD inputs closer to the uniform distribution:
\begin{align}
\setlength{\abovedisplayskip}{0.05cm}
\setlength{\belowdisplayskip}{0.01cm}
\mathcal{L}\!=\mathbb{E}_{\left(\mathbf{x_{ood}}\right) \sim \mathcal{D}}[-H(p_{\theta}(y|x_{ood}))]
\vspace{-0.9cm}
\end{align}
where $H$ is the Shannon entropy of the predicted distribution. $p_{\theta}(y|x_{ood})$ is the predicted distribution of the input OOD utterance $x_{ood}$.

\noindent\textbf{Bound}(\newcite{Liu2020EnergybasedOD}) uses a regularization loss defined in terms of energy to further widen the energy gap:
\vspace{-0.2cm}
\begin{align}
\setlength{\abovedisplayskip}{0.05cm}
\setlength{\belowdisplayskip}{0.01cm}
\begin{split}
\mathcal{L}\!=\!\mathbb{E}_{\left(\mathbf{x}_{\text {ind}}\right) \sim \mathcal{D}}\max \!(0, E(\mathbf{x}_{\text{ind}})\!-\text {m}_{\text{ind}}))^2 \\
\!+\mathbb{E}_{\left(\mathbf{x}_{\text {ood}}\right) \sim \mathcal{D}}\max \!(0, \text {m}_{\text{ood}} - E(\mathbf{x}_{\text{ood}})))^2
\label{eq_loss0}
\vspace{-0.9cm}
\end{split}
\end{align}
where $E$ is the energy score of IND or OOD samples in the train set. This learning objective using two squared hinge loss with two hyper-parameters  $m_{ind}$ and $m_{ood}$. Note that \textbf{Bound} aims at OOD image classification and replies on two independent energy bounds. Instead, our proposed Margin constructs a contrastive energy margin between IND intents and OOD intents to better disentangle energy distributions.

\section{Implementation Details}
We use the public pre-trained 300 dimensions GloVe embeddings \cite{pennington2014glove}\footnote{https://github.com/stanfordnlp/GloVe} to embed tokens. We use a two-layer BiLSTM as a feature extractor and set the dimension of hidden states to 128. The dropout value is fixed at 0.5. We use Adam optimizer \cite{kingma2014adam} to train our model. We set the learning rate to 1E-03. In the training stage, we use standard cross-entropy loss for unsupervised OOD and cross-entropy+energy-guided margin loss for supervised OOD. Besides, in supervised OOD scenario, we employ restriction-oriented random sampling. Specifically, we guarantee that IND and OOD samples are both included in each batch to facilitate calculation of margin loss. We both set the training epoch up to 200 with a early stop of patience 15. For our proposed energy-guided margin loss, we set the margin $m$ to 19.0 and the temperature $T$ to 0.8. We use the best OOD F1 scores on the validation set to calculate the threshold adaptively. Each result of the experiments is tested 5 times under the same setting and gets the average value. The training stage of our models lasts about 2 minutes for unsupervised OOD and 4 minutes for supervised OOD both on a single Tesla T4 GPU(16 GB of memory). The average value of the trainable model parameters is 3.05M. We will release our code after blind review.

\section{A Theoretical Proof of Energy Score vs Softmax Score}
\label{proof}
In this section, we give a theoretical proof of why energy score outperforms softmax score. Supposing we get the output logits from the intent classifier, we represent MSP as follows:
\begin{equation}
\begin{aligned}
log \ \mathbf{MSP}(logits) &=  log \ max\ \mathit{softmax}(logits) \\
    &= log\ max \frac{exp(logits_{i})}{\sum _{i}exp(logits_{i})} \\
    &= log\ \frac{exp\ max(logits)}{\sum _{i}exp(logits_{i})} \\
    &= max(logits) - log\ sum\ exp(logits) 
\end{aligned}
\end{equation}
where $logits_{i}$ represents the $i$-th value in the vector $logits$. Recap the energy definetion:
\begin{equation}
E(\mathbf{x} ; f)=-T \cdot \log \sum_{i}^{K} e^{f_{i}(\mathbf{x}) / T}
\label{eq3}
\end{equation}
Here we set $T$ to 1. Therefore, we get the following equation:
\begin{equation}
    log \ \mathbf{MSP}(logits)= \underbrace{max(logits)}_{regularization\ item} + \mathbf{Energy}(logits)
\end{equation}
If the output logits get a high max value, then $max(logits)$ performs as a regularization item to avoid energy score increasing. Therefore, energy score can better mitigate the overconfidence issue than softmax score.

\end{document}